\documentclass[letterpaper]{article} 
\usepackage{aaai25}  
\usepackage{times}  
\usepackage{helvet}  
\usepackage{courier}  
\usepackage[hyphens]{url}  
\usepackage{graphicx} 
\usepackage{natbib}  
\usepackage{caption} 
\usepackage{algorithm}
\usepackage{algorithmic}
\usepackage{multirow}
\usepackage{amssymb}
\usepackage{color}
\usepackage{amsmath}
\usepackage{booktabs}
\usepackage{newfloat}
\usepackage{listings}
\DeclareCaptionStyle{ruled}{labelfont=normalfont,labelsep=colon,strut=off} 
\floatstyle{ruled}
\newfloat{listing}{tb}{lst}{}
\floatname{listing}{Listing}
\nocopyright 

\title{END\textsuperscript{2}: Robust Dual-Decoder Watermarking Framework Against Non-Differentiable Distortions}
\author{
    Nan Sun\textsuperscript{\rm 1},
    Han Fang\textsuperscript{\rm 2},
    Yuxing Lu\textsuperscript{\rm 3},
    Chengxin Zhao\textsuperscript{\rm 1},
    Hefei Ling\textsuperscript{\rm 1}\footnote{Corresponding Author}
}
\affiliations{
    \textsuperscript{\rm 1} Huazhong University of Science and Technology\\
    \textsuperscript{\rm 2} National University of Singapore\\
    \textsuperscript{\rm 3} Peking University

    \{sunnan, chengxinzhao, lhfei\}@hust.edu.cn, fanghan@nus.edu.sg, yxlu0613@gmail.com
}

\begin{document}

\maketitle

\begin{abstract}
DNN-based watermarking methods have rapidly advanced, with the ``Encoder-Noise Layer-Decoder'' (END) framework being the most widely used. To ensure end-to-end training, the noise layer in the framework must be differentiable. However, real-world distortions are often non-differentiable, leading to challenges in end-to-end training. 
Existing solutions only treat the distortion perturbation as additive noise, which does not fully integrate the effect of distortion in training.
To better incorporate non-differentiable distortions into training, we propose a novel dual-decoder architecture (\textbf{END\textsuperscript{2}}). Unlike conventional END architecture, our method employs two structurally identical decoders: the Teacher Decoder, processing pure watermarked images, and the Student Decoder, handling distortion-perturbed images. 
The gradient is backpropagated only through the Teacher Decoder branch to optimize the encoder thus bypassing the problem of non-differentiability. To ensure resistance to arbitrary distortions, we enforce alignment of the two decoders' feature representations by maximizing the cosine similarity between their intermediate vectors on a hypersphere. 
Extensive experiments demonstrate that our scheme outperforms state-of-the-art algorithms under various non-differentiable distortions. Moreover, even without the differentiability constraint, our method surpasses baselines with a differentiable noise layer. Our approach is effective and easily implementable across all END architectures, enhancing practicality and generalizability.
\end{abstract}

\section{Introduction}

\begin{figure}[ht]
\centering
\includegraphics[width=1\columnwidth]{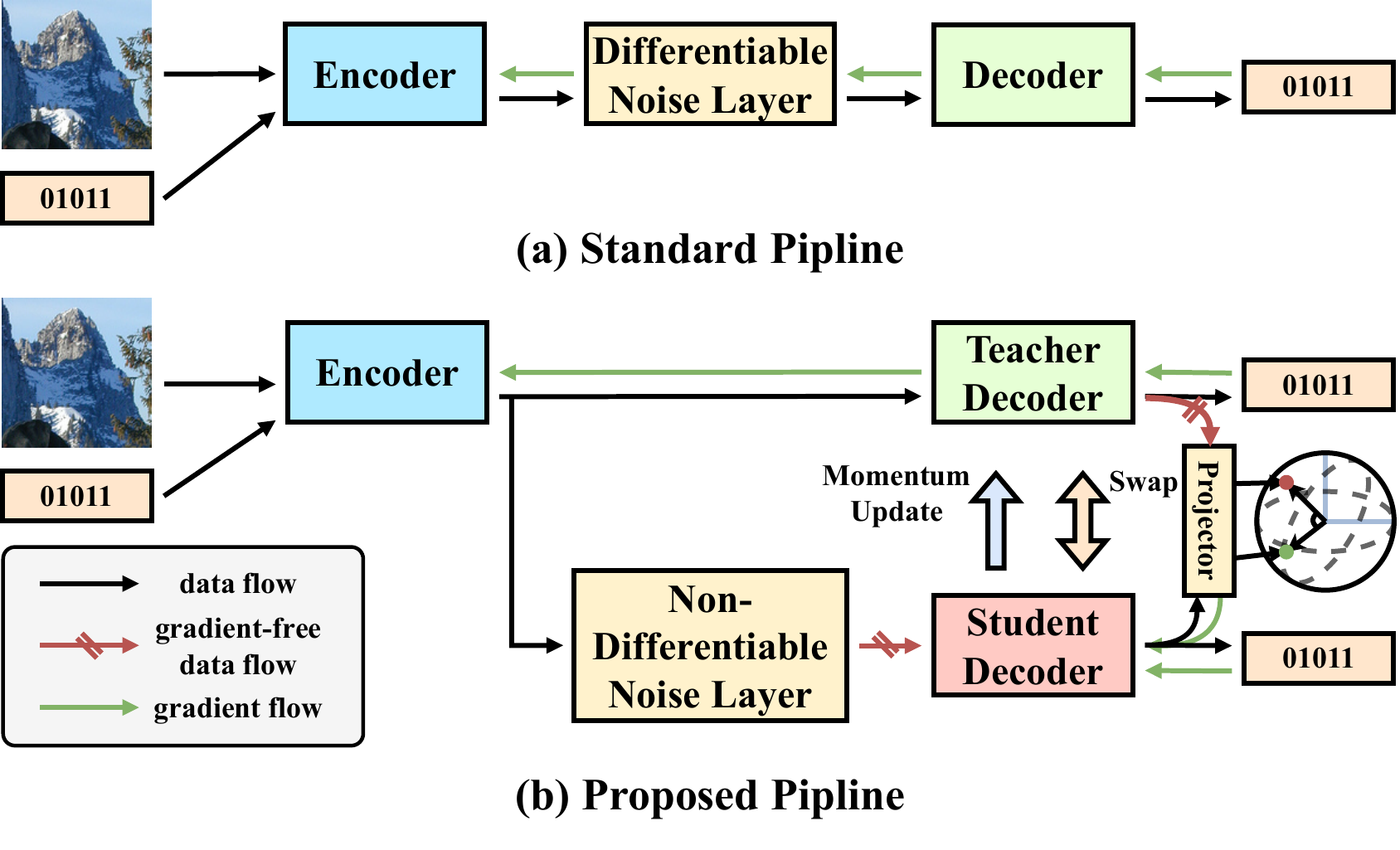}
\caption{The structures for different methods. (a) The standard END structure, which requires a differentiable noise layer to maintain the joint optimization of the model. (b) Our proposed END\textsuperscript{2} structure. Green arrows represent the direction of propagation of the gradient in backpropagation and the red arrow represents that the process is gradient-free.}
\label{fig:pipline}
\end{figure}
With the rapid growth of the Internet, accessing vast digital media resources has become easy, but this also increases the risk of unauthorized use, making protection and ownership verification crucial.
In order to verify ownership, digital watermarking technology was first introduced \cite{413536}. Such techniques have been widely studied in image \cite{hsu1999hidden, hernandez2000dct, bi2007robust}, video \cite{cox1997secure, langelaar2000watermarking, wang2006blind} and audio \cite{swanson1998robust, haitsma2000audio, bassia2001robust}. In recent years, DNN-based image watermarking \cite{kandi2017exploring, ahmadi2020redmark, zhu2018hidden, tancik2020stegastamp} has developed rapidly, with a series of models based on the ``Encoder-Noise Layer-Decoder'' (END) framework being the most widely used. The END framework employs an autoencoder-like architecture which contains an encoder, a noise layer, and a decoder, as shown in Figure \ref{fig:pipline} (a). The encoder embeds the watermark information into the cover image, while the decoder extracts the hidden information from the image. The noise layer enhances the model's robustness by applying various distortions.
To enable end-to-end joint training, the noise layer must be differentiable. However, real-world distortions
are typically non-differentiable (such as JPEG compression and photo filters in third party apps). This limitation significantly reduces the realistic effectiveness of existing models. Therefore, addressing the training for non-differentiable distortions is a crucial step toward developing a practical watermarking framework.

Current methods addressing non-differentiable distortions typically use mathematical modeling to create differentiable approximations \cite{zhu2018hidden, tancik2020stegastamp, li2024screen}, which fail to accurately replicate real-world distortions and are ineffective against black-box distortions with unknown implementations.
TDSL \cite{liu2019novel} uses a two-stage training strategy, training the model without noise initially and fine-tuning only the decoder with real noise later. However, the lack of joint training between the encoder and decoder reduces robustness to actual noise effects.
In addition, Forward ASL \cite{zhang2021towards} treats all noise as additive and considers only lossless watermarked images during backpropagation.
But treating all distortion as additive noise is too idealized to realistically reflect the effect of distortion on the model.

The above methods fail to effectively incorporate the impact of real-world noise during the training process, resulting in a gap between the model's learned behavior and the actual distortions encountered.
To address this problem, we propose a novel end-to-end dual-decoder architecture (\textbf{END\textsuperscript{2}}). The Figure \ref{fig:pipline} (b) illustrates our proposed architecture, which includes an encoder, a noise layer that does not require differentiability, and two decoders. As shown in the figure, in order to bypass the non-differentiable restriction, We utilize the Teacher Decoder branch to propagate the gradient of the clean encoded image directly to the encoder, facilitating the coordinated optimization of both the encoder and decoder. 
To effectively incorporate non-differentiable distortions and be robust against them, we introduce a Student Decoder, which shares the same structure as the Teacher Decoder but is specifically designed to process distorted images. By maximizing the cosine similarity between the feature vectors of the two decoders in the latent space, the model aligns the representations of distorted and clean images, thereby improving its robustness to non-differentiable distortions. 

In addition, we propose using swapping learning and momentum updating strategies to allow the two decoders to supervise each other, which ensures consistent feature representation, enhancing robustness to distortions.

The main contributions of this paper are as follows:
\begin{itemize}
    \item We propose END\textsuperscript{2}, a novel dual-decoder framework that effectively addresses the challenges posed by arbitrary non-differentiable distortions.
    \item Our approach is not only perfectly compatible with differentiable noise layers but also exhibits strong architectural generalization, making it applicable to various END architectures.
    \item Extensive experiments demonstrate that our method outperforms SOTA methods, exhibiting strong robustness to conventional non-differentiable noise as well as black-box distortions.
\end{itemize}

\section{Related Work}
\textbf{Deep Learning for Image Watermarking.} Recent advances in deep learning have significantly impacted the field of digital image watermarking, harnessing the powerful feature extraction capabilities of neural networks to improve robustness and visual quality. HiDDeN \cite{zhu2018hidden} pioneeres an end-to-end DNN-based watermarking framework with an autoencoder-like architecture, which has a profound impact on subsequent models. 
ReDMark\cite{ahmadi2020redmark} extends this approach by adopting a residual structure for the encoder and adjusting the intensity of the watermark pattern by an intensity factor. 
StegaStamp \cite{tancik2020stegastamp} focuses on print-shooting robustness, simulating the print-shooting process with several differentiable operations within the noise layer. The idea of distortion simulation has influenced the following researches. RIHOOP \cite{jia2020rihoop} adds a differentiable distortion network to simulate the distortion caused by camera imaging during the training process so that the watermark information is not affected by the camera. Another research \cite{li2024screen} goes a step further by considering the effect of grayscale deviation on the screen in the cross-media screen-camera process, thus constructing a more realistic distortion layer to improve the model performance. 

All of the above approaches maintain the robustness of the model by constructing a differentiable noise layer to approximate the perturbations of real-world images. This approach requires accurate modelling of the distortion, which is often difficult to achieve for complex distortions in the real world. Therefore, this approach is not generalisable.

\textbf{Image watermarking free from differentiable noise layer. } A number of approaches have been proposed to address the limitations of the differentiable noise layer. 
TDSL \cite{liu2019novel} separates the training process of the encoder and the decoder, thereby eliminating the differentiability constraint of the noise layer. In TDSL, the model is initially trained jointly without noise in the first stage, followed by training the decoder alone under real noise. However, this two-stage training strategy has an obvious problem: the encoder and decoder can only be optimized to their respective optimal solutions, but not to the global optimal solution, and it is difficult to ensure that the encoder will not be overfitted during the one-stage training process, thus resulting in the information being completely corrupted after the distortion. For this reason Forward ASL \cite{zhang2021towards} reintegrates the training process into an end-to-end architecture and treats all noise as additive noise, and considers only lossless watermarked images in the backpropagation without considering the effect of noise on the gradient. This approach is quite clean and simple, but since it does not consider the effect of noise on the backpropagation, this makes the gradient information obtained by the encoder may be too different from the true gradient information, which results in performance degradation.

TDSL's two-stage training strategy results in insufficient joint training between the encoder and decoder, and Forward ASL's approach oversimplifies by treating arbitrary distortions as additive noise. Both methods fail to effectively incorporate real non-differentiable distortions into end-to-end training. In our \textbf{Experiments}, we demonstrate that these limitations hinder both methods' ability to resist specific types of distortion.
For this reason, we need to find a new way to to jointly train the encoder and decoder end-to-end, while at the same time making the Decoder robust enough to various non-differentiable distortions. In addition, We hope that this approach is effective and general enough to be compatible with most watermarking frameworks.
\section{Methodology}
\subsection{Preliminary}
Before delving into the specific details of our method, we first outline the basic assumptions and premises underlying our approach.
DNN-based Image Watermarking often rely on differentiable approximations of distortions, which limits their effectiveness against real-world non-differentiable distortions. Therefore, it is essential to develop a watermarking solution capable of effectively handling any form of non-differentiable distortion.
The effectiveness of differentiable noise layers is typically attributed to data enhancement, with Forward ASL\cite{zhang2021towards} showing that distortions primarily impact forward propagation. Based on this, we suggest that the Encoder can bypass the noise layer and adaptively refine the embedding strategy by receiving gradients from the robust Decoder alone to solve the non-differentiable problem. In our experiments, we demonstrate this by analyzing residuals under various noise attacks. Therefore, the key challenge lies in effectively training a robust decoder.
Inspired by self-supervised contrastive learning \cite{grill2020bootstrap, caron2021emerging, zbontar2021barlow}, we align the feature vectors of distorted and clean watermarked images. This alignment ensures that if the clean image's features are correctly decoded, the distorted image's features will also be decoded correctly. Unlike self-supervised learning, our approach has a clear downstream goal (extracting watermark information), so we avoid issues with intermediate features collapsing into constant values and losing semantic information.

\subsection{Regular Pipline}
First we illustrate the pipeline of the general framework, which typically follows an END architecture in standard watermarking networks.

As shown in Figure \ref{fig:pipline} (a), it contains three main modules: Image Encoder $f_{\theta}$, Message Decoder $g_{\delta}$, and differentiable noise layer $\phi$, where $\theta,\delta$ are the parameters of the encoder and decoder, respectively. Given an image set $\mathcal{D}$, an image $x \in \mathbb{R}^{3\times h \times w}$ sampled uniformly from $\mathcal{D}$, and a binary watermark message $m \in \{0,1\}^n$ of length $n$. The watermarked image is denoted as $\hat{x}$. The image after noise attack is $\tilde{x}$, and the extracted watermark information by the decoder is $\hat{m}$. the watermark embedding process can be described as follows:
\begin{equation}
    \hat{x}=f_{\theta}(x,m)
\end{equation}
The simulation of a noise attack can be represented as follows:
\begin{equation}
    \tilde{x} = \phi(\hat{x})
\end{equation}
The process of decoding the watermark information can be represented as:
\begin{equation}
    \hat{m}=g_{\delta}(\tilde{x})
\end{equation}
Our goal is to simultaneously optimize $\theta,\delta$ such that the following equation holds:
\begin{equation}
    \arg\min_{\theta,\delta} \mathrm{E}_{x \sim \mathcal{D}, m \sim \{0, 1\}^n} \left [ \left \| \hat{x}-x \right \| + \lambda \left \| \hat{m} - m \right \| \right ]
\end{equation}
where $\lambda$ is the hyperparameter used to balance visual quality and decoding accuracy. From the above equation, it can be seen that $\tilde{x}$ is very critical for $g_\delta$. Because if $\tilde{x}$ itself does not carry information $m$, the decoder is unlikely to be better than a random selection no matter how much $\delta$ is optimised. Therefore $\lambda \left \| \hat{m} -m \right \|$ has to constrain the optimisation of not only $\delta$ but also $\theta$, which requires that the noise layer $\phi$ has to be differentiable, allowing the gradient to propagate from the decoder to the encoder. This limitation results in deep learning based watermarking frameworks not being able to maintain good robustness in the face of black-box noise and non-differentiable noise.

\subsection{END\textsuperscript{2} Pipline}
Our pipline is shown in Figure \ref{fig:pipline} (b). To illustrate, we split the decoder in more detail. Let $g_\delta(\tilde{x})=\gamma(\xi(\tilde{x}))$, $z\triangleq \xi(\tilde{x})$, where $\xi(\cdot)$ represents the feature extraction of the input image $\tilde{x}$ to obtain the feature vector $z\in \mathbb{R}^d$. The $d$ stands for the dimension of the latent space. The $\gamma(\cdot)$ then represents the prediction of the final watermark information based on the feature vector, which consists of a linear layer. In contrast to the regular pipline we employ two decoders with the same structure but different roles in the training process, called Teacher Decoder \textbf{TD} and Student Decoder \textbf{SD}. Since they have the same structure, we use $g_{t}$ and $g_{s}$ to differentiate them in the following. 

Our proposed method incorporates three additional components beyond the standard framework: (1) a dual-decoder for information extraction, (2) a feature alignment loss $\mathcal{L}_{s,t}$ into the primary loss function $\mathcal{L}$, and (3) the swapping learning and momentum updating strategy.

TD receives clean watermarked images to extract information and performs end-to-end optimization to ensure high quality watermark embedding and extraction. Instead SD receives the watermarked image with non-differentiable distortion directly. It serves to imitate the TD by aligning the feature vectors in the latent space, making the model robust to various distortions.

In addition, we have a projection layer $\varphi$ shared by TD and SD, whose role is to project the feature vectors of both decoders into the same high-dimensional space. It is implemented by a bias-free linear layer. And the non-differentiable noise layer $\psi$, which is used to implement various non-differentiable distortion operations. During experiments, in order to ensure its non-differentiable nature, we use the stop gradient technique $sg[\cdot]$ to simulate the non-differentiable process. 

\begin{algorithm}[htb]
\caption{END\textsuperscript{2} Pipeline}
\label{alg:algorithm}
\begin{algorithmic}[1] 
\REQUIRE Image-Message pairs $(x, m)$ from dataset, Encoder $f_{\theta}$, Teacher Decoder $g_t$ and Parameters $\delta_t$, Student Decoder $g_s$ and Parameters $\delta_s$, Distortion Function $\psi$, Momentum Coefficient $\tau$, Optimizer $\rm op$, Swapping interval $k$
\ENSURE Optimized Encoder and Decoders

\FOR{$i, (x, m)$ \textbf{in} enumerate(dataset)}
    \STATE $\hat{x} \gets f_{\theta}(x, m)$ \textcolor[rgb]{0.25,0.5,0.5}{\# Encode image with watermark}
    \STATE $\tilde{x}\gets sg[\psi(\hat{x})]$ \textcolor[rgb]{0.25,0.5,0.5}{\# Apply distortion attack}
    \STATE
    \STATE \textcolor[rgb]{0.25,0.5,0.5}{\# extract message and feature from encoded images}
    \STATE $\hat{m}_t, z_t \gets g_t(\hat{x})$ \textcolor[rgb]{0.25,0.5,0.5}{\# Teacher decoder}
    \STATE $\hat{m}_s, z_s \gets g_s(\tilde{x})$ \textcolor[rgb]{0.25,0.5,0.5}{\# Student decoder}
    \STATE
    \STATE $\mathcal{L} \gets \mathcal{L}_{s,t} + \mathcal{L}_{msg} + \mathcal{L}_{quality}$ \textcolor[rgb]{0.25,0.5,0.5}{\# Calculate total loss}
    \STATE $\mathcal{L}$.backward() \textcolor[rgb]{0.25,0.5,0.5}{\# Backpropagate}
    \STATE $\rm op$.step() \textcolor[rgb]{0.25,0.5,0.5}{\# Update model parameters}
    \STATE
    \STATE\textcolor[rgb]{0.25,0.5,0.5}{\# Momentum updating strategy}
    \STATE $\delta_t \gets \tau \delta_t + (1 - \tau) \delta_s$ 
    \STATE \textcolor[rgb]{0.25,0.5,0.5}{\# Swapping Learning strategy}
    \IF{$(i + 1)\mod k = 0$}
        \STATE $g_t, g_s \gets g_s, g_t$ 
    \ENDIF
\ENDFOR
\end{algorithmic}
\end{algorithm}

The algorithm \ref{alg:algorithm} shows the training flow of our method. 
Compared to the general END model process, our method adds one more Student Decoder and introduces momentum updating and swapping learning after training step. Furthermore, our approach remains compatible with differentiable noise layers without modification, enabling seamless integration with all END models.

\subsection{Feature Alignment in Latent Space}
In order for SD to learn how to deal robustly with a variety of non-differentiable distortions, we project the intermediate features of the two decoders onto a hypersphere and minimize the angle between their feature vectors. 

First extract the feature vectors:
\begin{equation}
    \begin{aligned}z_t&=\xi_t(\hat{x})\\ z_s&=\xi_s(sg[\psi(\hat{x})])\end{aligned}
\end{equation}

The $sg[\cdot]$ here is only added to ensure that the differentiable distortion also maintains its non-differentiable character in experiments, and can be removed in deployment. 

Afterwards we project both feature vectors onto the same hypersphere via the projection layer $\varphi$:
\begin{equation}
    \begin{aligned}\bar{z}_t&=\frac{\varphi(z_t)}{\left \| \varphi(z_t)\right \|_2} \\ \bar{z}_s&= \frac{\varphi(z_s)}{\left \| \varphi(z_s)\right \|_2} \end{aligned}
\end{equation}

We also refer to the above operation as the projection operation $\mu(\cdot)$.

Finally we use cosine similarity to bring $\bar{z}_s$ closer to $\bar{z}_t$, which we call feature alignment loss:
\begin{equation}
    \mathcal{L}_{s,t}=2-2\frac{\left \langle \bar{z}_s,sg[\bar{z}_t] \right \rangle }{\left\|\bar{z}_s\right\|_2\cdot\left\|sg[\bar{z}_t]\right\|_2}\triangleq\left\|\bar{z}_s-st[\bar{z}_t]\right\|_2^2
\end{equation}
Here we only let $z_s$ be close to $z_t$, hence the need for gradient truncation function $sg[\cdot]$. Considering it from another perspective, we are essentially clustering in the latent space of watermarked information, and $\bar{z}_t$ is the centre of our clustering, and we keep all the feature vectors of the noisy image with the same information close to this centre.

\subsection{Total Loss Function}
Apart from the the feature alignment loss $\mathcal{L}_{s,t}$, our model includes the message loss $\mathcal{L}_{msg}$, aimed at ensuring the model can effectively extract the embedded information, and the quality loss $\mathcal{L}_{quality}$, which is essential for maintaining the visual quality of the watermarked images.

The message loss can be expressed by the following equation:
$$\mathcal{L}_{msg}=\left \|\hat{m}_t - m \right \|_2 + \left \| \hat{m}_s - m\right \|_2$$

The quality loss can be represented as follows:
$$\mathcal{L}_{quality}=\left \| \hat{x} -x \right \|_2$$

The total loss can be expressed as:
$$\mathcal{L}=\lambda_1\mathcal{L}_{s,t}+\lambda_2\mathcal{L}_{msg}+\lambda_3\mathcal{L}_{quality}$$
where $\lambda_1,\lambda_2,\lambda_3$ are three hyperparameters used to balance different losses, defaulting to $0.01, 8, 5$.

\subsection{Swapping Learning and Momentum Updating}
Since the TD only processes clean watermarked images, it cannot directly learn a decoding method robust to distortions. Consequently, the feature alignment loss will let the SD to emulate TD’s representation ability, potentially leading to performance degradation. To ensure the consistency of representational capabilities between the two decoders, we employ swapping learning and momentum updating strategies. These approaches effectively enhance the robustness of our model.


\textbf{Swapping Learning.} To ensure that the TD experiences distortions, we introduce a straightforward approach: swapping the two decoders after each $k$ training batches, with $k$ defaulting to $1$. What's more, the feature alignment loss ensures that, after the swap, the encoded features of the original TD can close to those of the original SD. This mutual supervision between the decoders facilitates the generation of consistent representations.

\textbf{Momentum Updating.} 
This strategy enables the TD to learn from the non-differentiable noise processed by the SD, thereby producing features better suited for decoding noisy images. Additionally, constant momentum updating helps ensure that both decoders gradually converge to similar parameters, maintaining consistent decoding capabilities. The process is illustrated as follows:

\begin{equation}
    \begin{aligned}\delta_s &\longleftarrow \rm{optimizer}(\delta_s,\nabla \mathcal{L}_{s,t}, \eta )\\\delta_t&\longleftarrow \tau\delta_t+(1-\tau)\delta_s\end{aligned}
\end{equation}
where $\tau$ is the weight factor and $\eta$ is the learning rate, defaulting to $0.999$ and $8e^{-4}$.

In Section \textbf{Ablation Study}, we will show that both strategies are effective in improving the performance of the model. And when the two are combined, they can produce even better results.


\begin{figure*}[ht]
\centering
\includegraphics[width=1\textwidth]{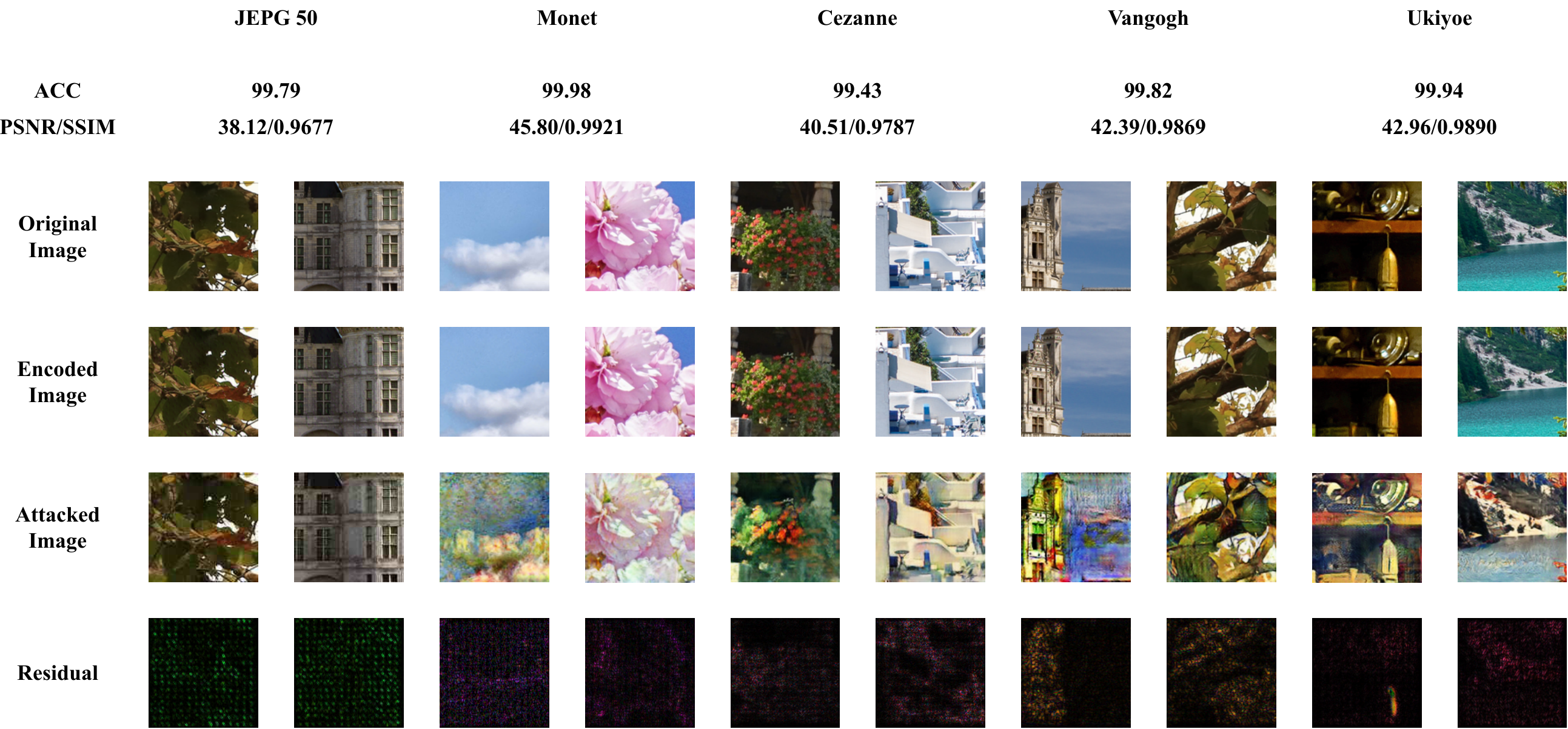}
\caption{Results of the invisibility and robustness of our model under real JPEG compression and four style transfer distortions. The second row depicts the original image, while the third row shows the embedded watermarked image. The fourth row illustrates the watermarked image after being subjected to non-differentiable distortion. The final row shows the residual image, which is the difference between watermarked images and original images, magnified by a factor of $10$ for enhanced visibility.}
\label{fig:blackbox}
\end{figure*}
\section{Experiments}
\subsection{Implementation Details}
Our primary interest is to explore the performance of END\textsuperscript{2} under fully non-differentiable distortion. In addition we also explore whether our approach is more competitive compared to other watermarking models that require a differential noise layer. Since our approach is a generic framework, we do not concern ourselves with the specific implementations of the Encoder and Decoder. Therefore, we employ the model structure in  MBRS \cite{jia2021mbrs}. The model is trained on DIV2K\cite{agustsson2017ntire}. Specifically, we randomly select a block of size $128\times 128$ from the training set as the cover image, and randomly sample a bit stream of length $30$ to use as watermarking information for embedding. To evaluate the model's generalization, we also randomly select $2000$ images from the COCO \cite{lin2014microsoft} and ImageNet \cite{deng2009imagenet} datasets for testing. We utilize the Adam optimizer with a fixed learning rate of $8e^{-4}$ for training. The batch size is set to $32$, and the model is trained for a total of $5000$ epochs on a NVIDIA RTX 3090 GPU.

\textbf{Metrics.} We use average bit accuracy (ACC) to evaluate the decoding accuracy of the model. For visual quality assessment, we use the peak signal-to-noise ratio (PSNR) and 
 structural similarity (SSIM) as metrics. 

\textbf{Baselines.} We compare against TDSL \cite{liu2019novel} and Forward ASL \cite{zhang2021towards} for non-differentiable distortion. Additionally, to demonstrate the superiority of our method, we compare it with three classical models HiDDeN\cite{zhu2018hidden}, StegaStamp\cite{tancik2020stegastamp} and MBRS \cite{jia2021mbrs} which need differentiable Noiser. 

\subsection{Comparison with Previous Methods}
\begin{table}[t]
\renewcommand\arraystretch{1.2}
\resizebox{\columnwidth}{!}{

\begin{tabular}{c|cccccc}
\toprule
                                                          & END\textsuperscript{2}   & TDSL   & Forward ASL & MBRS   & HiDDeN & StegaStamp \\ \midrule
\begin{tabular}[c]{@{}c@{}}Non-Diff\\ Noiser\end{tabular} & \checkmark    & \checkmark    & \checkmark         &        &        &            \\
ACC                                                       & \textbf{94.55}  & 94.10  & 92.37       & 89.17  & 87.86  & 88.90      \\
PSNR                                                      & \textbf{45.62}  & 36.32  & 33.48       & 40.64  & 32.62  & 35.34      \\
SSIM                                                      & \textbf{0.9897} & 0.9726 & 0.9054      & 0.9796 & 0.8486 & 88.10      \\ \bottomrule
\end{tabular}

}
\caption{Average decoding accuracy and visual quality of the different methods under the random distortion.}
\label{tab:combine}
\end{table}
In this section, we compare our method with two SOTA models \cite{liu2019novel, zhang2021towards} that also address non-differentiable distortion. At the same time we choose two classical models \cite{zhu2018hidden, tancik2020stegastamp} that require a differentiable noise layer to illustrate that even without the need for a differentiable noise layer our approach is still competitive. In addition, since our model maintains the same structure as MBRS \cite{jia2021mbrs}, we trained the MBRS model under differentiable conditions for comparison.

We use a combined noise to simulate complex distortion conditions in the real world. The noise layer consists of Roate, Crop, Translate, Scale, Shear, Dropout, Cropout, Color transformation, JPEG compression, Gaussian Filter, Gaussian Noise. 

\begin{table*}[]
    \centering
    
\renewcommand\arraystretch{1.2}
\resizebox{\linewidth}{!}{

\begin{tabular}{ccccccccccc}
\hline
Dataset                   & Method      & Identity     & \begin{tabular}[c]{@{}c@{}}Gaussian Filter\\ ($\sigma=2$)\end{tabular} & \begin{tabular}[c]{@{}c@{}}JPEG\\ ($Q=50$)\end{tabular} & \begin{tabular}[c]{@{}c@{}}Crop\\ ($p=0.1$)\end{tabular} & \begin{tabular}[c]{@{}c@{}}Dropout\\ ($p=0.5$)\end{tabular} & \begin{tabular}[c]{@{}c@{}}Rotate\\ ($deg=10$)\end{tabular} & \begin{tabular}[c]{@{}c@{}}Translate\\ ($dis=0.05$)\end{tabular} & \begin{tabular}[c]{@{}c@{}}Scale\\ ($f=0.65$)\end{tabular} & \begin{tabular}[c]{@{}c@{}}Gaussian Noise\\ ($std=0.01$)\end{tabular} \\ \hline
\multirow{6}{*}{DIV2K}    & Proposed    & \textbf{100} & \textbf{99.46}                                                         & \textbf{95.67}                                          & 93.21                                                    & \textbf{100}                                                & 98.26                                                       & \textbf{99.76}                                                   & \textbf{94.37}                                             & \textbf{100}                                                          \\
                          & TDSL        & 98.92        & 60.18                                                                  & 55.21                                                   & 64.87                                                    & 99.56                                                       & \textbf{99.85}                                              & 81.67                                                            & 88.71                                                      & 98.89                                                                 \\
                          & Forward ASL & 99.99        & 83.68                                                                  & 90.21                                                   & 92.43                                                    & 99.23                                                       & 92.48                                                       & 89.38                                                            & 70.63                                                      & 99.99                                                                 \\
                          & MBRS        & 98.16        & 95.31                                                                  & 86.43                                                   & \textbf{94.24}                                           & 98.59                                                       & 95.22                                                       & 94.70                                                            & 91.78                                                      & 98.14                                                                 \\
                          & HiDDeN      & 89.56        & 59.66                                                                  & 53.45                                                   & 87.04                                                    & 71.50                                                       & 82.62                                                       & 82.65                                                            & 68.59                                                      & 88.31                                                                 \\
                          & StegaStamp  & 90.94        & 89.92                                                                  & 84.55                                                   & 88.61                                                    & 77.53                                                       & 86.60                                                       & 86.54                                                            & 84.10                                                      & 90.76                                                                 \\ \hline
\multirow{6}{*}{COCO}     & Proposed    & \textbf{100} & \textbf{99.22}                                                         & \textbf{94.89}                                          & \textbf{93.08}                                           & \textbf{99.99}                                              & 98.21                                                       & \textbf{99.77}                                                   & \textbf{93.77}                                             & \textbf{100}                                                          \\
                          & TDSL        & 99.77        & 58.32                                                                  & 53.94                                                   & 63.52                                                    & 96.72                                                       & \textbf{99.75}                                              & 80.36                                                            & 86.33                                                      & 99.63                                                                 \\
                          & Forward ASL & 99.98        & 85.57                                                                  & 92.05                                                   & 88.33                                                    & 98.46                                                       & 92.02                                                       & 88.98                                                            & 69.74                                                      & 99.94                                                                 \\
                          & MBRS        & 96.99        & 94.92                                                                  & 90.59                                                   & 90.74                                                    & 90.39                                                       & 90.79                                                       & 91.21                                                            & 91.01                                                      & 91.69                                                                 \\
                          & HiDDeN      & 86.26        & 73.11                                                                  & 66.49                                                   & 70.92                                                    & 70.44                                                       & 71.87                                                       & 72.76                                                            & 72.00                                                      & 73.50                                                                 \\
                          & StegaStamp  & 89.58        & 89.33                                                                  & 87.56                                                   & 87.41                                                    & 84.91                                                       & 85.18                                                       & 85.20                                                            & 85.01                                                      & 85.59                                                                 \\ \hline
\multirow{6}{*}{ImageNet} & Proposed    & \textbf{100} & \textbf{99.33}                                                         & \textbf{93.71}                                          & \textbf{94.13}                                           & \textbf{99.99}                                              & \textbf{98.33}                                              & \textbf{99.79}                                                   & \textbf{93.68}                                             & \textbf{100}                                                          \\
                          & TDSL        & 99.59        & 59.73                                                                  & 54.06                                                   & 64.60                                                    & 97.70                                                       & 98.28                                                       & 81.20                                                            & 86.59                                                      & 99.38                                                                 \\
                          & Forward ASL & 99.94        & 86.83                                                                  & 91.94                                                   & 89.08                                                    & 98.64                                                       & 91.11                                                       & 88.40                                                            & 69.70                                                      & 99.97                                                                 \\
                          & MBRS        & 96.42        & 93.85                                                                  & 89.59                                                   & 90.00                                                    & 95.19                                                       & 90.19                                                       & 90.67                                                            & 90.36                                                      & 90.97                                                                 \\
                          & HiDDeN      & 86.24        & 72.54                                                                  & 66.17                                                   & 70.49                                                    & 69.97                                                       & 71.37                                                       & 72.21                                                            & 71.35                                                      & 72.80                                                                 \\
                          & StegaStamp  & 88.99        & 88.33                                                                  & 86.63                                                   & 86.23                                                    & 83.71                                                       & 83.89                                                       & 83.90                                                            & 83.49                                                      & 84.00                                                                 \\ \hline
\end{tabular}

}

    \caption{Benchmark comparisons on robustness against different distortions. We trained the models only on DIV2K and tested them on three different datasets. By adjusting the embedding strength, the visual quality of all models was maintained at PSNR=35.}
    \label{tab:single}
\end{table*}

To quantitatively analyze the relationship between visual quality and decoding accuracy, we measure the performance of different models under random noise attacks. Table \ref{tab:combine} shows the performance under random distortion. Our approach, TDSL and Forward ASL do not require a differentiable noise layer, so we truncate the gradient of the distorted image in the actual training. 
Our method significantly outperforms the baselines. Although TDSL achieves similar decoding accuracy, its redundant embedding strategy results in a PSNR nearly $10$ points lower than ours. Notably, our approach even surpasses MBRS trained with a differentiable noise layer, demonstrating that such a layer is non-essential.

For a comprehensive evaluation, we test the decoding accuracy of our trained model across three datasets using a variety of distortions: Identity, Gaussian Filter ($\sigma=2$), JPEG compression ($Q=50$), Crop ($p=0.1$), Dropout ($p=0.5$), Rotate ($deg=10$), Translate ($dis=0.05$), Scale ($f=0.65$) and Gaussian Noise($std=0.01$). Table \ref{tab:single} presents the experimental results. Our method maintains a decoding accuracy of over $90\%$ under all noise attacks, demonstrating its robustness against a wide range of distortions. TDSL struggles to resist Gaussian Filter, JPEG Compression, and Crop attacks, while Forward ASL performs poorly under Scale attacks. This suggests that both methods are less robust to certain types of distortions, despite their claims of handling arbitrary distortions.

\subsection{Performance under Non-Differentiable Distortion}


To further demonstrate the robustness of our method under real non-differentiable distortions, we select real JPEG Compression and four style transfer black-box distortions for testing. For JPEG compression, we utilized the PIL package in Python, while the four style transfer distortions are generated by a pre-trained CycleGAN \cite{zhu2017unpaired} model. 

\textbf{Real JPEG Compression.} We train our model using only real JPEG Compression with a quality factor of $Q=50$ and then test it under various quality factors. To be fair, we fix the PSNR at $38$. Notably, the model trained with the combined noise layer in the previous subsection is expected to demonstrate some level of generalization performance. Therefore, we consider it as a baseline to evaluate its robustness against real-world black-box noise, thereby underscoring the necessity of training against black-box distortions.

Table \ref{tab:jpeg} presents the results of our experiments. Our proposed method consistently outperforms most baseline models, proving the validity of our approach. Moreover, it is evident that the model pre-trained with a combined noise layer does not perform well on real JPEG compression, even though simulated JPEG compression was included in the training process. This discrepancy indicates a significant gap between the simulated noise layer and real-world noise. Therefore, it is crucial to develop a method that can robustly handle arbitrary black-box distortions encountered in real-world scenarios.
\begin{table}[t]
\centering
\renewcommand\arraystretch{1.2}
\resizebox{0.9\linewidth}{!}{

\begin{tabular}{c|ccccc}
\toprule
            & \multicolumn{5}{c}{Real JPEG Compression} \\ \cmidrule{2-6} 
            & 50     & 40     & 30     & 20     & 10    \\ \midrule
Proposed    & \textbf{99.95}  & 94.39  & \textbf{83.41}  & \textbf{68.22}  & \textbf{55.20} \\
Pre-trained & 69.02  & 64.45  & 60.53  & 55.58  & 51.49 \\
TDSL        & 74.58  & 70.05  & 63.50  & 57.32  & 52.19 \\
Forward ASL & 99.76  & \textbf{96.39}  & 80.43  & 60.83  & 51.41 \\ \bottomrule
\end{tabular}
}
\caption{Result of real JPEG compression. }
\label{tab:jpeg}
\end{table}


\textbf{Black-box Style Transfer Distortions.} Testing with style transfer distortions is essential. Because it simulates complex, non-differentiable transformations that occur in the real world. Unlike traditional noise or compression, style transfer introduces variations in texture, color, and structure, challenging the model with intricate distortions. To validate our method, we selected four styles of migration distortions (Monet, Cezanne, Vangogh and Ukiyoe) . 

As illustrated in Figure \ref{fig:blackbox}, our method maintains a decoding accuracy above $99\%$ across various distortions, while also preserving high visual quality. Furthermore, to illustrate that our approach allows the Encoder to perceive different distortions, we draw the JPEG Compression with quality factor $Q=50$ also in Figure \ref{fig:blackbox}. 
The residual images reveal that the encoder adopts distinct encoding strategies for different types of distortions. This indicates that our approach enables the Encoder to recognize patterns in non-differentiable distortions and apply a more robust coding strategy.

\begin{figure}[tb]
\centering
\includegraphics[width=\columnwidth]{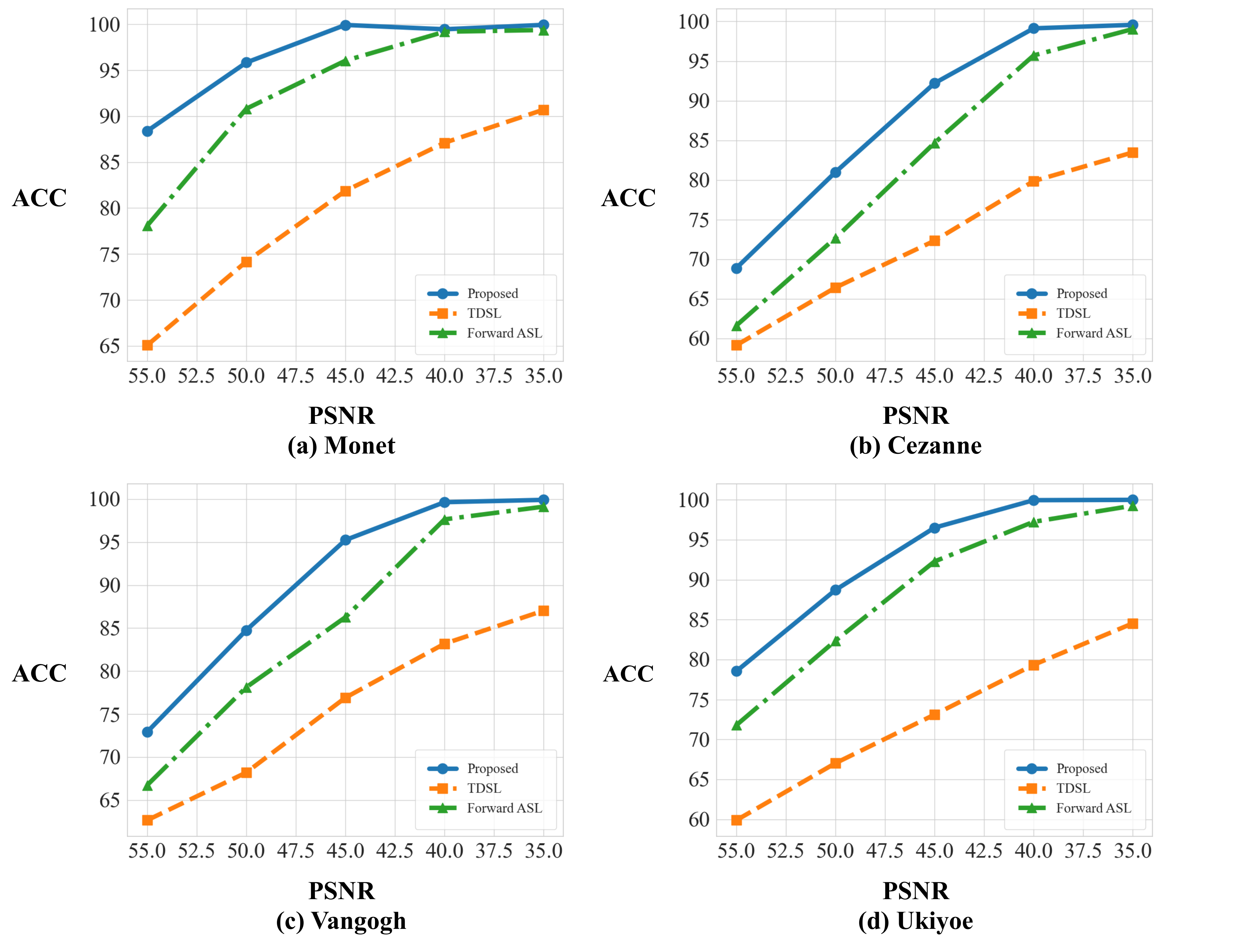}
\caption{The performance of different models, in terms of average decoding accuracy, under four style transfer distortions is evaluated as PSNR varies.}
\label{fig:style}
\end{figure}

In addition, we evaluated the robustness of each model against the four style transfer distortions by varying the watermark embedding strength. As shown in Figure \ref{fig:style}, our method consistently outperforms the baseline at various watermark strengths and achieves nearly $95\%$ decoding accuracy at a PSNR of $45$. These results demonstrate that our approach is robust to unknown black-box distortions.

\subsection{Ablation Study} \label{sec:ab}
In this section, we conduct the ablation experiments to better illustrated the proposed architecture. We are primarily interest in the effectiveness of the momentum updating and swapping learning strategies, and how different feature alignment losses affect the performance of the model.

\begin{table}[t]
\centering
\renewcommand\arraystretch{1.2}
\resizebox{0.75\columnwidth}{!}{

\begin{tabular}{l|lll}
\toprule
         & ACC   & PSNR  & SSIM   \\ \midrule
None     & 78.93 & 45.04 & 0.9889 \\
FA       & 86.79 & \textbf{47.89} & \textbf{0.9921} \\
FA+MU    & 92.23 & 39.26 & 0.9620 \\
FA+SL    & 93.96 & 43.68 & 0.9867 \\
FA+MU+SL & \textbf{94.55} & 45.62 & 0.9896 \\ \bottomrule
\end{tabular}
}
\caption{The performance of the model under different structures. Where FA, MU, SL stand for feature alignment loss, momentum updating strategy and swapping learning strategy respectively.}
\label{tab:ab}
\end{table}



\textbf{Importance of Different Compositions.} Table \ref{tab:ab} presents the ablation study results of our method under various configurations. The first row represents the outcomes without applying any specific strategies, while the subsequent rows illustrate the results obtained using different combinations of our proposed strategies. The experiments reveal that training without strategies results in a final decoding accuracy of approximately $78\%$. Introducing the feature alignment loss increases this accuracy by about $8\%$, demonstrating that aligning features in the latent space can effectively enhance the model's robustness. Furthermore, employing either the MU or SL strategies individually leads to additional improvements in robustness, albeit with a reduction in visual quality. When both strategies are applied simultaneously, the model achieves a peak decoding accuracy of approximately $95\%$ while maintaining a PSNR of $45$.

\begin{table}[t]
\centering
\renewcommand\arraystretch{1.2}
\resizebox{0.75\linewidth}{!}{

\begin{tabular}{l|lll}
\toprule
         & ACC   & PSNR  & SSIM   \\ \midrule
Proposed & 94.55 & \textbf{45.62} & \textbf{0.9896} \\
MSE      & \textbf{94.70} & 42.92 & 0.9837       \\
DINO     & 91.39 & 43.23      & 0.9855       \\ \bottomrule
\end{tabular}
}
\caption{Impact of different feature alignment losses on model performance.}
\label{tab:loss}
\end{table}

\textbf{Impact of Feature Alignment Loss.} Furthermore, we explored the effect of different feature alignment losses on model performance by comparing the Mean Squared Error (MSE) loss and the feature alignment method mentioned in DINO \cite{caron2021emerging}. Table \ref{tab:loss} presents the results of our experiments. Experiments indicate that although the direct use of MSE loss achieves a comparable ACC to our method, it results in a visual quality drop of about $3$ points. In contrast, using DINO leads to a decrease in ACC. However, our method consistently maintains performance, showing that the choice of alignment loss does not significantly impact the overall effectiveness.

\section{Conclusion}

In this paper, we presented a novel dual-decoder architecture (END\textsuperscript{2}) to address the challenges posed by non-differentiable distortions in deep learning-based image watermarking. By aligning feature vectors between a Teacher Decoder for lossless images and a Student Decoder for distorted images, our approach overcomes the limitations of traditional END architectures that rely on differentiable noise layers. Experiments show that our method outperforms state-of-the-art algorithms in various distortion scenarios and can be easily integrated into existing END frameworks. Our findings highlight the potential of END\textsuperscript{2} as a robust solution for real-world image watermarking challenges.

\section{Acknowledgments}
This work was supported in part by the Natural Science Foundation of China under Grant 61972169,62372203 and 62302186, in part by the National key research and development program of China(2022YFB2601802), in part by the Major Scientific and Technological Project of Hubei Province (2022BAA046, 2022BAA042), in part by the Knowledge Innovation Program of Wuhan-Basic Research, in part by China Postdoctoral Science Foundation 2022M711251.

\bigskip


\end{document}